\documentclass{amia}

\usepackage{graphicx}        
\usepackage{xcolor}        

\usepackage{amsmath,amssymb,amsfonts}  
\usepackage{booktabs}        
\usepackage{multirow}        
\usepackage{tabularx}       
\usepackage{algorithm}      
\usepackage{algpseudocode}   
\usepackage{float}    
\usepackage{subcaption} 

\usepackage{pifont} 

\newcommand{\cmark}{\ding{51}}  
\newcommand{\xmark}{\ding{55}}  
\begin{document}

\title{A Framework for Cross-Domain Generalization in Coronary Artery Calcium Scoring Across Gated and Non-Gated Computed Tomography}

\author{Mahmut S. Gokmen$^1$, Moneera N. Haque$^3$, Steve W. Leung$^4$, Caroline N. Leach$^1$, Seth Parker$^2$, Stephen B. Hobbs $^4$, Vincent L. Sorrell$^3$, W. Brent Seales$^2$, V. K. Cody Bumgardner$^1$}

\institutes{
    $^1$ Center for Applied AI, University of Kentucky, Lexington, KY\\
    $^2$ EduceLab, University of Kentucky, Lexington, KY \\
    $^3$ Department of Medicine, Division of Cardiovascular Medicine and the Gill Heart and Vascular Institute, University of Kentucky, Lexington, KY\\
    $^4$ Department of Radiology, College of Medicine, University of Kentucky, Lexington, KY
}
\maketitle

\section*{Abstract}

\textit{Coronary artery calcium (CAC) scoring is a key predictor of cardiovascular risk, but relies on ECG-gated CT scans, restricting their use to specialized cardiac imaging settings. We introduce an automated framework for CAC detection and lesion-specific Agatston scoring that works across gated and non-gated CT scans. At its core is \textbf{CARD-ViT}, a self-supervised Vision Transformer trained exclusively on gated CT data using DINO. Without any non-gated training data, our framework achieves 0.707 accuracy and Cohen’s $\kappa$ of 0.528 on the Stanford non-gated dataset, matching models trained directly on non-gated scans. On gated test sets, the framework achieves 0.910 accuracy with Cohen’s $\kappa$ scores of 0.871 and 0.874 across independent datasets, demonstrating risk stratification. Results demonstrate the feasibility of cross-domain CAC scoring from gated to non-gated domains, supporting scalable cardiovascular screening in routine chest imaging without additional scans or annotations.}

\section{Introduction} Cardiovascular disease (CVD), primarily driven by coronary artery disease (CAD), is the leading global cause of mortality \cite{CVD_2024, NCHS_2024}. A primary manifestation of CAD is the buildup of hardened plaque in the heart's arteries, a condition known as Coronary Artery Calcification (CAC). The total amount of this calcified plaque is a powerful and independent predictor of future cardiovascular events, such as heart attacks \cite{Hecht2015-gd}. Consequently, CAC scoring, a non-invasive quantification of this plaque burden from non-contrast gated CT scans, is a widely recommended risk stratification tool \cite{Knuuti2020-xr,Oudkerk2008-zt}. This score, typically calculated with the Agatston method \cite{Agatston1990-to}, directly informs clinical decisions: scores $\geq 100$ often prompt statin therapy, while a score of 0 can justify deferring treatment \cite{Tsao2023-ae}.
ECG-gated CT scans synchronize image acquisition with the cardiac cycle to minimize motion artifacts and enable more precise quantification of coronary calcium. However, non-gated CT scans are performed approximately 10 times more frequently for evaluating pulmonary conditions and therefore represent a substantial missed opportunity for opportunistic cardiovascular screening without requiring additional imaging or radiation exposure \cite{Raygor2023-rv,Liu2022-vc}. Manual or semi-automatic analysis remains time-consuming (5-10 minutes per case) and operator-dependent, motivating the development of automated deep learning methods \cite{Eng2021-da,Van_Assen2021-mj,Takahashi2023-ah}.

Recent automated scoring methods have used both gated and non-gated CTs~\cite{CAC_review_2, gated_vs_nongated, Zeleznik2021-ab}. However, a significant domain shift between these modalities, motion artifacts, often requires complex learning strategies ~\cite{CAC_review_1,CAC_review_2,Eng2021-da}. Some propose domain adaptation techniques, using gated data to "prompt" scoring on non-gated scans \cite{combatin}. While effective, these methods still require access to both data domains during training, which poses a practical limitation given the scarcity of large-scale, well-curated non-gated, labeled datasets ~\cite{gated_vs_nongated, CAC_review_1,CAC_review_2}. Consequently, many models are evaluated on limited test cohorts, potentially yielding misleading conclusions about real-world generalizability.

This paper introduces an end-to-end framework for automated coronary artery calcium detection, pixel-level segmentation, and lesion-specific Agatston scoring that achieves cross-domain generalization from gated to non-gated CT scans. We bypass mixed-domain training \cite{combatin} and test the hypothesis that a model trained exclusively on high-quality gated data can learn robust representations that generalize to unseen non-gated scans. At its core is \textbf{CARD-ViT}, a Vision Transformer backbone trained using self-supervised DINO \cite{dinov2, dino_register}. This pipeline, developed without non-gated supervision, generalizes effectively, performing reliable lesion-specific Agatston scoring on both gated and non-gated scans, and includes deployment via MONAI Label and OHIF \cite{lesion_cac,lesion_cac_2,monailabel, OHIF,orthancserver}.
\section{Methods}
The framework follows a three-stage pipeline: (1) self-supervised representation learning for feature extraction, (2) lesion-specific segmentation and Agatston-based scoring, and (3) deployed for clinical evaluation using the MONAI Label server integrated with the OHIF Viewer interface.
\begin{figure*}[htbp]
\centering
\includegraphics[width=0.9\linewidth]{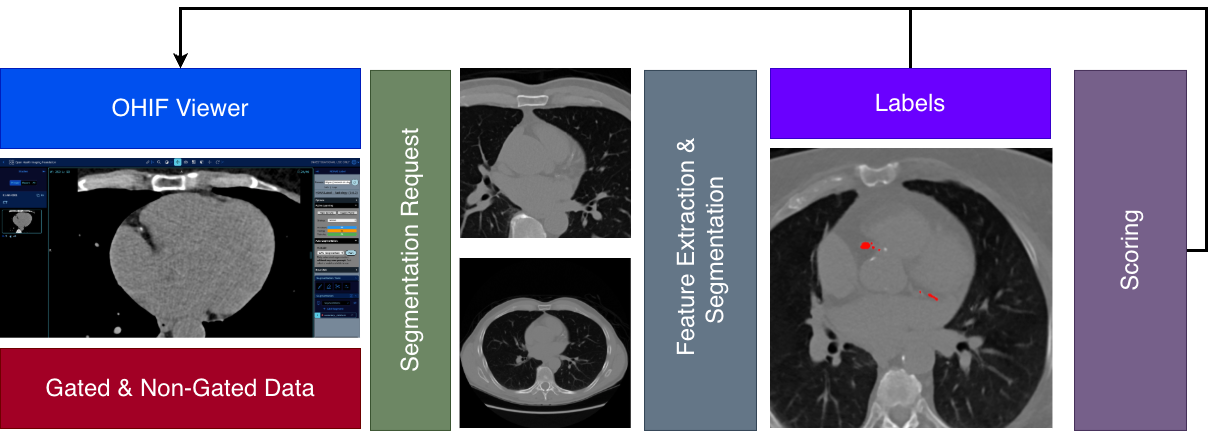}
\caption{Overview of the automated CAC scoring architecture for coronary artery calcium detection and quantification.}
\label{fig:overall_arch}
\end{figure*} 

\textbf{Feature Extraction with CARD-ViT Model.}  
We introduce CARD-ViT, a Vision Transformer (ViT) backbone pre-trained with DINO (self-DIstillation with NO labels) and adapted for medical CT imaging. ViTs encode CT slices into compact embeddings that capture complex spatial relationships and positional context, enabling detailed analysis of calcification patterns. The self-supervised training approach allows the model to learn robust and generalizable features from large amounts of unlabeled CT scans, making it effective on both gated and non-gated scans without additional manual annotations. The RGB-based patch embedding layer is modified to accept single-channel CT inputs by averaging pretrained RGB weights across channels, preserving compatibility with grayscale data. CARD-ViT processes input CT slices ($512 \times 512$ pixels) by dividing them into non-overlapping $16 \times 16$ patches, resulting in $32 \times 32 = 1024$ patch tokens. These patches are linearly projected into a $768$-dimensional embedding space and processed through multiple transformer encoder layers alongside a \texttt{[CLS]} token and four register tokens (auxiliary learnable embeddings that refine spatial attention patterns). The model extracts three feature types from the final transformer layer: (1) \textit{patch embeddings} (768 channels, 32×32) encoding local content, (2) \textit{attention maps} (12 channels) derived from \texttt{[CLS]}-to-patch attention, and (3) \textit{register token features} (4 channels) from register-to-patch attention. These features are concatenated channel-wise to form a unified $784$-channel representation.
\begin{figure*}[htbp]
    \centering
    \begin{subfigure}[b]{0.45\textwidth}
        \centering
        \includegraphics[width=\linewidth]{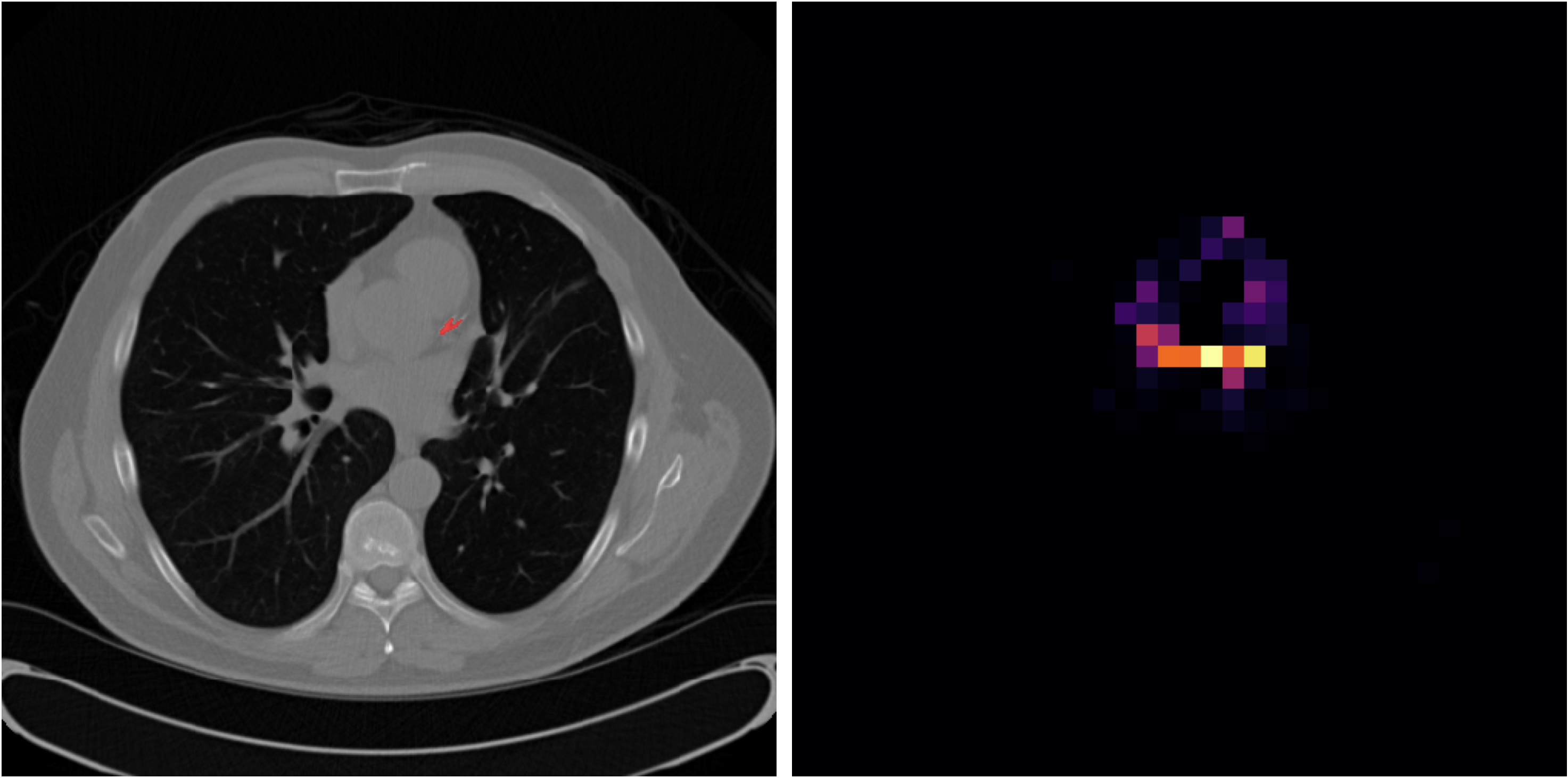}
        \caption{Non-gated CT}
        \label{fig:non_gated_attention}
    \end{subfigure} 
    \begin{subfigure}[b]{0.45\textwidth}
        \centering
        \includegraphics[width=\linewidth]{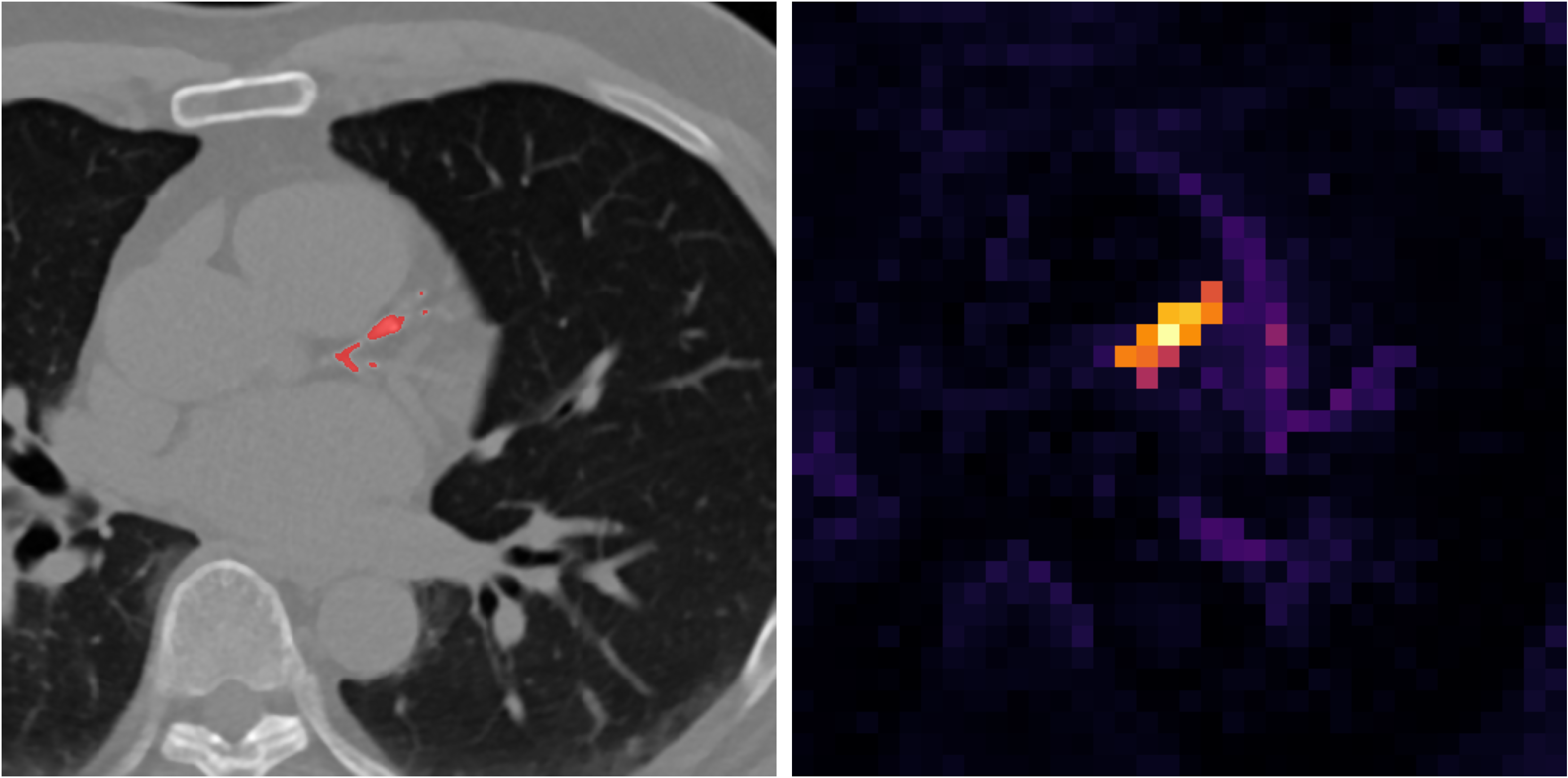}
        \caption{Gated CT}
        \label{fig:gated_attention}
    \end{subfigure}
\caption{Attention maps from CARD-ViT highlighting calcified regions in non-gated (a) and gated (b) CT scans. Each pair shows a raw CT slice (left) and the corresponding PCA-transformed attention visualization (right) from the self-supervised ViT backbone.}

\label{fig:attention_visualizations}
\end{figure*}

\begin{figure*}[htbp]
\centering
\includegraphics[width=0.95\linewidth]{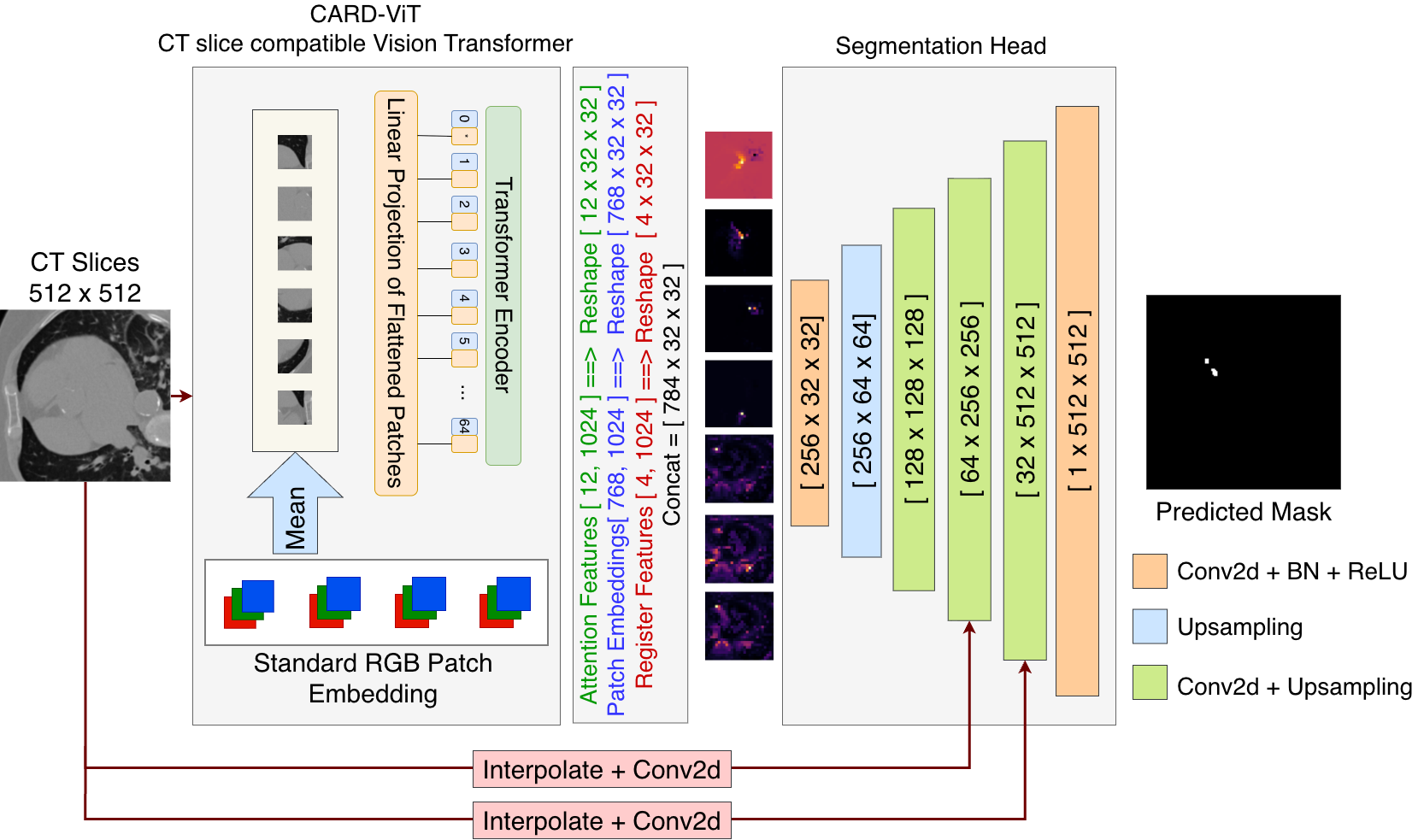}
\caption{CARD-ViT segmentation pipeline: ViT feature extraction, $784$-channel feature fusion, decoder upsampling with selective skip connections, and final binary mask output.}
\label{fig:segmentation_process}
\end{figure*}

Although ViTs lack inherently high-resolution outputs due to patch-based tokenization, their attention mechanisms provide interpretable insight into model focus. As shown in Figure~\ref{fig:attention_visualizations}, the attention maps are derived directly from the self-supervised CARD-ViT backbone and highlight relevant anatomical structures such as coronary calcifications. This confirms that the model learns to focus on clinically important regions without using any explicit supervision. This multi-scale feature representation enables segmentation head to reconstruct precise segmentation masks as detailed in Figure~\ref{fig:segmentation_process}.

\textbf{Segmentation and CAC Scoring.}
Following feature extraction, the unified $784$-channel representation generated from CARD-ViT is processed by a custom decoder for pixel-level coronary calcification segmentation~\cite{vit_unet}. As shown in Figure~\ref{fig:segmentation_process}, an initial convolutional block reduces the dimensionality of the $784$ to $256$ channels at resolution $32 \times 32$ to preserve captured calcification features. The decoder then upsamples through four stages ($32 \rightarrow 64 \rightarrow 128 \rightarrow 256 \rightarrow 512$) via bilinear interpolation and convolutional refinement to generate calcification masks with the same size of input CT slice ($512\times512$). To restore fine anatomical details lost during decoding, selective skip connections from the input CT slice are integrated at $256 \times 256$ and $512 \times 512$ stages. A final $1 \times 1$ convolutional head outputs a \textbf{single-channel binary mask} highlighting all calcified regions, rather than producing separate masks for each artery (e.g., LM, LAD, CX, RCA). This design improves robustness across both gated and non-gated CT scans and ensures methodological compatibility for direct comparison of our approach with prior works such as AI-CAC~\cite{nejm_model}, which also employed single-channel masks for non-gated data. 
Consolidating artery-specific masks also mitigates class imbalance issues common in multi-channel outputs while preserving spatial fidelity.

The resulting segmentation mask is passed to a lesion-specific adaptation of the Agatston scoring method in clinical standard~\cite{CAC_scoring_1}. Connected-component analysis identifies individual calcified lesions, and for each lesion, the physical area (in mm$^2$) is combined with the maximum Hounsfield Unit (HU) intensity to compute a weighted Agatston score. Lesion-level scores are then summed across all slices to obtain the total patient CAC score. This formulation better captures morphological and spatial characteristics of calcified plaque compared to slice-wise scoring, and improves consistency with clinical risk models~\cite{lesion_cac, lesion_sup11}. Moreover, prior studies have linked lesion-specific scores above 200 with adverse outcomes such as stent underexpansion~\cite{stent_cac_2, clinic_cac_23}, further validating the utility of this approach. Finally, our scoring method is fully compatible with external benchmarks like AI-CAC~\cite{nejm_model}, enabling standardized evaluation across datasets and clinical workflows.

\textbf{Clinical Deployment with MONAI Label and OHIF Viewer.}
Our pipeline was deployed in a clinical evaluation environment leveraging three core components: (1) the OHIF Viewer (visualization), (2) an Orthanc server (DICOM storage), and (3) a locally-hosted AI model service running CARD-ViT. These components are integrated using the MONAI Label framework, which facilitates real-time inference and annotation feedback \cite{monailabel,orthancserver,OHIF}. Due to incompatibilities between Monai Label and Vision Transformer architectures arising from dependencies, we deployed a separate model inference endpoint to run CARD-ViT independently and communicate with the OHIF interface. This configuration allowed radiologists to review segmentation outputs, interactively refine results, and conduct real-time validation of model predictions.

\begin{figure*}[htbp]
\centering
\includegraphics[width=0.87\linewidth]{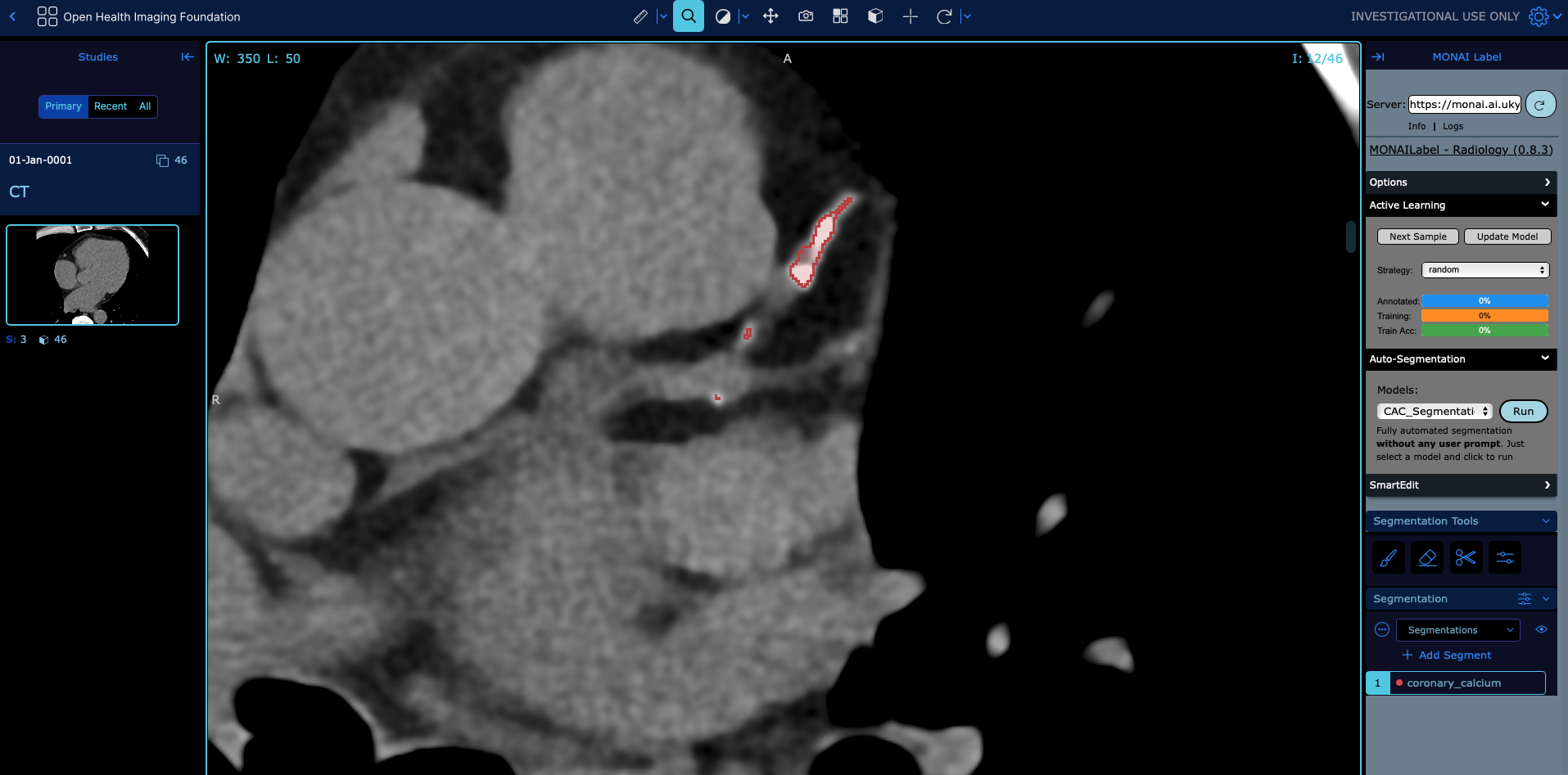}
\caption{Automated segmentation by CARD-ViT (red overlay) visualized in the OHIF Viewer via MONAI Label for real-time clinical review.}
\label{fig:OhifSS}
\end{figure*}

\section{Results}
To evaluate the effectiveness of our framework, we assessed its two key components, the \textbf{CARD-ViT} backbone and the \textbf{segmentation head}, on internal (Heartlens) and external (Stanford) test sets comprising gated and non-gated CT scans \cite{COCA_Dataset}. On gated scans, our framework achieved high segmentation and scoring accuracy, generalizing across institutions without site-specific tuning or retraining (Table \ref{tab:gated_performance}). For non-gated evaluation, we compared against \textbf{AI-CAC}~\cite{nejm_model}, a publicly available model trained specifically for CAC scoring on non-gated CTs. AI-CAC was selected as a clinically relevant comparator due to its peer-reviewed validation, widespread use, and architectural similarity, especially its use of \textbf{single-channel lesion segmentation} rather than vessel-specific masks, aligning with our design. This ensured a fair comparison under similar scoring strategies. Notably, although trained solely on gated CTs, our model achieved comparable classification performance to AI-CAC on non-gated scans. Unlike AI-CAC, which was trained directly on non-gated data, our approach required no domain-specific supervision—highlighting its potential to overcome data scarcity challenges while maintaining clinical accuracy in diverse real-world settings.

\textbf{Dataset.}  
The proposed framework was evaluated using two real-world clinical datasets: the \textbf{Heartlens} dataset  from the University of Kentucky, acquired through support of the Center for Clinical and Translational Science and the \textbf{Stanford Coronary Calcium} dataset~\cite{COCA_Dataset}. The Heartlens dataset contains only gated CT scans and was split into training (2,651 cases) and internal testing (469 cases) sets. \textbf{The model was trained exclusively on the Heartlens training set.} Patient-level Agatston scores were available for all cases, while pixel-level calcium annotations were provided for a subset of slices (4.3\% for training, 4.2\% for testing).  
The Stanford dataset served entirely as an external test set and was not used during training. It includes two independent cohorts: a \textbf{gated subset} of 443 CTs with full pixel-level annotations (used for segmentation evaluation and CAC scoring evaluation), and a \textbf{non-gated subset} of 205 CTs with patient-level scores but no segmentation labels (used solely for CAC scoring evaluation).  
Table~\ref{tab:dataset_complete} summarizes dataset composition, annotation coverage, CAC score distribution, and vessel-wise breakdown for all datasets. Vessel annotations showed imbalance across arteries: RCA (42.0\%), LAD (29.8\%), CX (21.7\%), and LM (6.5\%). The total vessel-annotated slice count (10,310) exceeds the total number of labeled slices (8,237), as some slices include multiple vessel labels.
\begin{table}[htbp]
\centering
\caption{Comprehensive dataset statistics. Top section shows dataset-level case/slice counts, annotation coverage, and CAC score distribution. Bottom section shows vessel-wise annotation breakdown for Heartlens gated cohort (all splits combined). "Labeled" indicates availability of pixel-level annotations. Percentages in case/slice columns indicate annotation coverage. "Score" indicates patient-level CAC score availability; "Label" indicates pixel-level segmentation label availability. \textbf{For the Stanford (Gated Test) cohort, the CAC score distribution was derived from the provided pixel-level ground-truth (GT) annotations, as patient-level scores were unavailable.}}
\label{tab:dataset_complete}
\resizebox{\textwidth}{!}{
\begin{tabular}{l rrr rrr cc cccc}
\toprule
& \multicolumn{3}{c}{\textbf{Case-Level}} & \multicolumn{3}{c}{\textbf{Slice-Level}} & \multicolumn{2}{c}{\textbf{Available}} & \multicolumn{4}{c}{\textbf{CAC Score Distribution}} \\
\cmidrule(lr){2-4} \cmidrule(lr){5-7} \cmidrule(lr){8-9} \cmidrule(lr){10-13}
\textbf{Dataset / Vessel} & Total & Labeled & \% & Total & Labeled & \% & Score & Label & 0--10 & 11--100 & 101--400 & 400+ \\
\midrule
\multicolumn{13}{l}{\textit{Dataset-Level Statistics}} \\
\midrule
Heartlens (Gated Train) & 2,651 & 727 & 27.4 & 162,584 & 7,028 & 4.3 & \cmark & \cmark & 1,220 & 504 & 456 & 471 \\
Heartlens (Gated Test)  & 469   & 129 & 27.5 & 28,873  & 1,209 & 4.2 & \cmark & \cmark & 216   & 89  & 81  & 83  \\
Stanford (Gated Test)   & 443   & 443 & 100.0 & 22,695  & 3,602 & 15.9 & \xmark* & \cmark & 77    & 140 & 103 & 123 \\
Stanford (Non-Gated)    & 205   & --  & --   & 9,222   & --    & --  & \cmark & \xmark & 105   & 42  & 32  & 27  \\
\midrule
\multicolumn{13}{l}{\textit{Vessel-Wise Annotations (Heartlens Gated: All Splits Combined)}} \\
\midrule
\quad LM  (Left Main)       & 317 & 317 & 100 & 10,310 & 666   & 6.5  & \cmark & \cmark & -- & -- & -- & -- \\
\quad LAD (Left Anterior)   & 753 & 753 & 100 & 10,310 & 3,072 & 29.8 & \cmark & \cmark & -- & -- & -- & -- \\
\quad CX  (Circumflex)      & 480 & 480 & 100 & 10,310 & 2,240 & 21.7 & \cmark & \cmark & -- & -- & -- & -- \\
\quad RCA (Right Coronary)  & 564 & 564 & 100 & 10,310 & 4,332 & 42.0 & \cmark & \cmark & -- & -- & -- & -- \\
\bottomrule
\end{tabular}
}
\end{table}

\textbf{Training Configuration.}
Self-supervised pretraining of the CARD-ViT backbone was conducted using the DINOv2-with-registers framework with standard contrastive loss~\cite{dinov2, dino_register}. Pretraining used the entire Heartlens training cohort (2,651 gated cases, 162,584 slices; see Table~\ref{tab:dataset_complete}), leveraging all slices (annotated or not) to maximize representation learning. Due to the small size (205 cases) and lack of pixel-level labels in the Stanford non-gated cohort, it was excluded from training to prevent domain imbalance and preserve cross-domain evaluation.  
Following pretraining, CARD-ViT weights were frozen, and a custom decoder was trained in a supervised fashion using only 7,028 annotated gated slices from the Heartlens training set. The decoder was optimized using a combination of Dice and Focal loss to address foreground-background class imbalance. Training ran for 20 epochs using Distributed Data Parallel (DDP) on two NVIDIA A6000 GPUs with batch size 16 per GPU and AdamW optimizer (1e-4 learning rate, 1e-5 weight decay). No data augmentation was applied, as frozen CARD-ViT features were sufficient for robust segmentation with minimal overfitting.

\textbf{Segmentation Performance.}
Quantitative segmentation performance was assessed on the gated test cohort, distinguishing between slices with and without ground-truth (GT) annotations. For slices containing GT annotations, standard pixel-level metrics (Dice, IoU, Precision, Recall) were computed to evaluate mask overlap (Table~\ref{tab:segmentation_results}). For the more numerous non-annotated slices, corresponding to regions without labeled calcification, the model’s predictions were examined for spurious activations. The model demonstrated high specificity, correctly predicting empty masks for the vast majority of these slices. This evaluation was limited to the gated cohorts, as pixel-level annotations were not available for the non-gated dataset. 
\begin{table}[htbp]
\centering
\caption{Segmentation performance on annotated slices from gated test sets. Metrics computed only on slices with ground-truth annotations. $N$ indicates the total number of test slices; metrics are averaged over annotated slices within each test set.}
\label{tab:segmentation_results}
\small
\setlength{\tabcolsep}{6pt}
\begin{tabular}{lccccc}
\toprule
\textbf{Dataset} & \textbf{N (Total Slices)} & \textbf{Dice} & \textbf{IoU} & \textbf{Precision} & \textbf{Recall} \\
\midrule
Heartlens Test & 28,873 & 0.865 & 0.857 & 0.869 & 0.867 \\
Stanford Test  & 22,695 & 0.839 & 0.829 & 0.844 & 0.845 \\
\bottomrule
\end{tabular}
\end{table}

On the internal Heartlens test set, the model achieved a Dice coefficient of 0.865 and IoU of 0.857, indicating substantial overlap between predicted segmentation masks and ground-truth labels. When evaluated on the external Stanford gated test set, performance remained robust with a Dice coefficient of 0.839. The modest performance gap between datasets is expected given domain shift from different scanner manufacturers, acquisition protocols, and patient populations. Nevertheless, these results confirm the model's ability to generalize effectively to unseen gated data from independent clinical sites without requiring site-specific fine-tuning.

\begin{figure*}[htbp]
\centering
\includegraphics[width=0.3\linewidth]{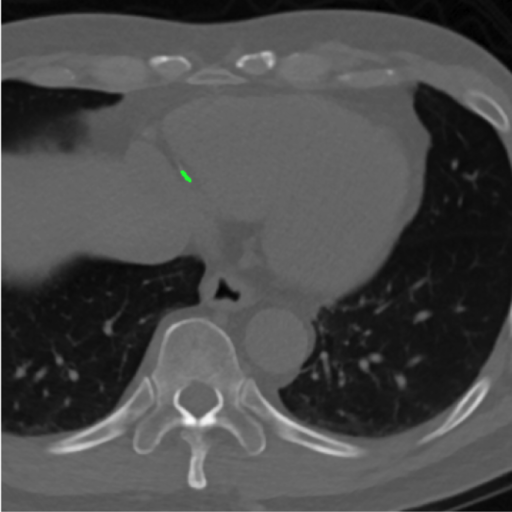}
\includegraphics[width=0.3\linewidth]{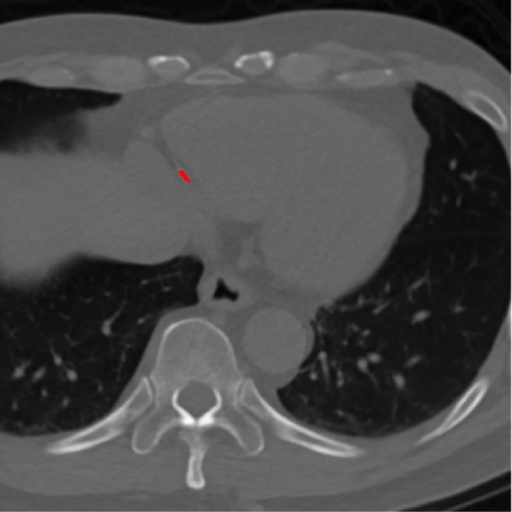}
\includegraphics[width=0.3\linewidth]{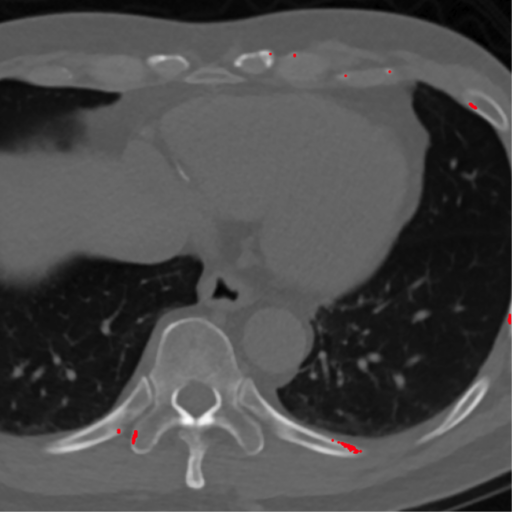}
\caption{Qualitative segmentation comparison on gated CT slices. \textbf{Left:} Ground-truth annotation overlay. \textbf{Middle:} CARD-ViT prediction showing accurate calcification localization. \textbf{Right:} AI-CAC prediction (trained on non-gated data) demonstrating domain mismatch with over-segmentation artifacts.}
\label{fig:segmentation_comparison}
\end{figure*}
Figure~\ref{fig:segmentation_comparison} presents qualitative segmentation results on representative gated CT slices. The first column shows ground-truth mask overlays, the second column displays predictions from CARD-ViT, and the third column includes segmentation output from the AI-CAC model for reference. It is important to note that AI-CAC was trained exclusively on non-gated CT data and exhibits significant domain mismatch when applied to gated scans, resulting in over-segmentation and false positives. Therefore, AI-CAC was excluded from quantitative evaluation on gated datasets, and its predictions are shown only for qualitative comparison to illustrate the importance of domain-matched training data.

\textbf{CAC Scoring Performance.}
CAC risk stratification was evaluated using lesion-specific Agatston scoring~\cite{lesion_cac}, enabling standardized comparison across gated and non-gated scans and with external benchmarks. Performance was assessed through sensitivity, specificity, positive/negative predictive values (PPV/NPV), F1-score, overall accuracy, and Cohen's $\kappa$ coefficient. Table~\ref{tab:cac_overall_summary} provides an overview of model performance across all test cohorts.

\begin{table}[htbp]
\centering
\caption{Summary of CAC scoring performance across gated and non-gated test cohorts. CARD-ViT was trained exclusively on gated data; non-gated evaluation represents cross-domain transfer. AI-CAC \cite{nejm_model} serves as a baseline trained directly on non-gated scans.}
\label{tab:cac_overall_summary}
\begin{tabular}{lccc}
\toprule
\textbf{Test Cohort} & \textbf{Model} & \textbf{Accuracy} & \textbf{Cohen's $\kappa$} \\
\midrule
Heartlens (Gated)    & CARD-ViT       & 0.910 & 0.871 \\
Stanford (Gated)     & CARD-ViT       & 0.910 & 0.874 \\
\midrule
Stanford (Non-Gated) & CARD-ViT       & 0.707 & 0.528 \\
Stanford (Non-Gated) & AI-CAC         & 0.707 & 0.542 \\
\bottomrule
\end{tabular}
\end{table}

\textit{Gated CT Performance.}
On gated test sets, CARD-ViT achieved robust risk classification with overall accuracies of 0.910 and Cohen's $\kappa$ values exceeding 0.87 on both Heartlens and Stanford cohorts, indicating strong agreement with ground truth. As shown in Table~\ref{tab:gated_performance}, per-category F1-scores ranged from 0.85 to 0.95 across all risk groups, with highest performance in extreme categories (0--10 and 400+). Confusion matrices reveal that misclassifications predominantly occurred between adjacent risk categories, a clinically acceptable pattern that minimally impacts treatment decisions. For instance, only 10 of 216 Heartlens cases in the 0--10 category were misclassified as 11--100, with negligible confusion to higher-risk groups.

\begin{table*}[htbp]
\centering
\caption{Gated CT performance on Heartlens and Stanford test sets. Top section shows per-category performance metrics; bottom section displays confusion matrices with ground truth in rows and predictions in columns.}
\label{tab:gated_performance}
\begin{tabular}{lccccc|ccccc}
\toprule
\multirow{2}{*}{\textbf{CAC Range}} &
\multicolumn{5}{c|}{\textbf{Heartlens}} &
\multicolumn{5}{c}{\textbf{Stanford}} \\
\cmidrule(lr){2-6} \cmidrule(lr){7-11}
& Sens & Spec & PPV & NPV & F1 & Sens & Spec & PPV & NPV & F1 \\
\midrule
0--10     & 0.93 & 0.99 & 0.98 & 0.96 & 0.95 & 0.82 & 0.97 & 0.94 & 0.89 & 0.88 \\
11--100   & 0.88 & 0.97 & 0.84 & 0.98 & 0.86 & 0.90 & 0.96 & 0.91 & 0.97 & 0.90 \\
101--400  & 0.86 & 0.97 & 0.84 & 0.98 & 0.85 & 0.91 & 0.97 & 0.85 & 0.98 & 0.88 \\
400+      & 0.95 & 0.99 & 0.91 & 0.99 & 0.94 & 0.97 & 0.98 & 0.92 & 0.99 & 0.95 \\
\midrule
\textbf{Overall (Acc/$\kappa$)} & \multicolumn{5}{c|}{\textbf{0.910 / 0.871}} & \multicolumn{5}{c}{\textbf{0.910 / 0.874}} \\
\midrule
\midrule
\multicolumn{11}{c}{\textit{Confusion Matrices (Ground Truth $\times$ Predicted)}} \\
\midrule
\textbf{Ground Truth} & \multicolumn{5}{c|}{\textbf{Heartlens Predicted}} & \multicolumn{5}{c}{\textbf{Stanford Predicted}} \\
\cmidrule(lr){2-6} \cmidrule(lr){7-11}
& 0--10 & 11--100 & 101--400 & 400+ & & 0--10 & 11--100 & 101--400 & 400+ & \\
\midrule
0--10     & 201 & 10 & 2  & 3  &  & 63  & 7   & 6   & 1  & \\
11--100   & 4   & 77 & 7  & 0  &  & 4   & 126 & 6   & 4  & \\
101--400  & 2   & 4  & 70 & 5  &  & 0   & 6   & 94  & 3  & \\
400+      & 0   & 1  & 4  & 78 &  & 0   & 0   & 4   & 119 & \\
\bottomrule
\end{tabular}
\end{table*}

\textit{Non-Gated CT Generalization.}
To assess cross-domain transferability, CARD-ViT was evaluated on the Stanford non-gated cohort without any fine-tuning or domain adaptation. Despite training exclusively on gated data, the model achieved 0.707 accuracy with Cohen's $\kappa$ of 0.528, matching the performance of AI-CAC~\cite{nejm_model}, which was trained directly on non-gated scans, demonstrating robust cross-domain generalization.
Performance analysis revealed complementary strengths between the two models. CARD-ViT demonstrated superior sensitivity for zero-calcium cases (0.933 vs. 0.876), effectively identifying patients without coronary calcification despite domain shift. However, it exhibited lower sensitivity in the high-risk category (0.593 vs. 0.778), suggesting that gated training data may underrepresent the morphological characteristics of severe calcifications in non-gated scans. Both models struggled with intermediate risk groups (11--400), where limited sample sizes (41--42 cases per category) and heterogeneous plaque distributions complicate accurate classification. Confusion matrices reveal that most errors involved underestimation in the 11--100 range, with 24 cases misclassified as 0--10 by CARD-ViT versus 21 by AI-CAC, a clinically significant pattern that could delay intervention for patients with emerging coronary disease.
\begin{table*}[htbp]
\centering
\caption{Cross-domain generalization performance on non-gated CT scans, comparing CARD-ViT (trained only on gated data) with AI-CAC (trained directly on non-gated data) on the Stanford non-gated test set. Top section shows per-category metrics; bottom section displays confusion matrices with ground truth in rows and predictions in columns.}
\label{tab:nongated_comparison}
\small
\begin{tabular}{l|ccccc|ccccc}
\toprule
\multirow{2}{*}{\textbf{CAC Range}} & \multicolumn{5}{c|}{\textbf{CARD-ViT (Gated $\rightarrow$ Non-Gated)}} & \multicolumn{5}{c}{\textbf{AI-CAC (Trained on Non-Gated)}} \\
\cmidrule(lr){2-6} \cmidrule(lr){7-11}
& Sens & Spec & PPV & NPV & F1 & Sens & Spec & PPV & NPV & F1 \\
\midrule
0--10     & 0.933 & 0.720 & 0.778 & 0.911 & 0.848 & 0.876 & 0.770 & 0.800 & 0.856 & 0.836 \\
11--100   & 0.415 & 0.896 & 0.500 & 0.860 & 0.453 & 0.463 & 0.890 & 0.514 & 0.869 & 0.487 \\
101--400  & 0.438 & 0.942 & 0.583 & 0.901 & 0.500 & 0.406 & 0.936 & 0.542 & 0.895 & 0.464 \\
400+      & 0.593 & 0.972 & 0.762 & 0.940 & 0.667 & 0.778 & 0.955 & 0.724 & 0.966 & 0.750 \\
\midrule
\textbf{Overall (Acc/$\kappa$)} & \multicolumn{5}{c|}{\textbf{0.707 / 0.528}} & \multicolumn{5}{c}{\textbf{0.707 / 0.542}} \\
\midrule
\midrule
\multicolumn{11}{c}{\textit{Confusion Matrices (Ground Truth $\times$ Predicted)}} \\
\midrule
\textbf{Ground Truth} & \multicolumn{5}{c|}{\textbf{CARD-ViT Predicted}} & \multicolumn{5}{c}{\textbf{AI-CAC Predicted}} \\
\cmidrule(lr){2-6} \cmidrule(lr){7-11}
& 0--10 & 11--100 & 101--400 & 400+ & & 0--10 & 11--100 & 101--400 & 400+ & \\
\midrule
0--10     & 98 & 5  & 2  & 0  &  & 92 & 6  & 5  & 2  & \\
11--100   & 24 & 17 & 0  & 0  &  & 21 & 19 & 1  & 0  & \\
101--400  & 2  & 11 & 14 & 5  &  & 2  & 11 & 13 & 6  & \\
400+      & 2  & 1  & 8  & 16 &  & 0  & 1  & 5  & 21 & \\
\bottomrule
\end{tabular}
\end{table*}

These results establish that self-supervised ViT representations learned from gated data effectively transfer to non-gated CT scans without domain-specific supervision. Minor score-level discrepancies are expected due to implementation-specific differences in Agatston scoring (e.g., lesion connectivity and peak-HU handling), and such offsets can disproportionately affect intermediate ranges (11--400) near category thresholds. The comparable $\kappa$ values (0.528 vs. 0.542) suggest that motion artifacts and acquisition variability in non-gated scans remain key challenges.

\textbf{Qualitative Clinical Validation.} 
To assess clinical relevance, clinical co-authors performed a qualitative review of segmentation predictions using the deployed OHIF viewer. Their feedback confirmed that the model distinguishes true coronary calcifications from common and clinically challenging confounders. It was specifically noted that the model successfully avoided segmenting adjacent non-coronary calcified structures, including the aortic root, mitral annular calcification, and nearby calcified lymph nodes. Furthermore, the model demonstrated robust performance in complex scenarios by correctly disregarding high-density artifacts such as pacemaker leads while still identifying true calcifications. Clinical co-authors also noted cases where the model detected subtle calcifications that could be overlooked during manual review.

The assessment also identified limitations including occasional false positives from adjacent aortic root calcification and rare false negatives on single slices. Cases involving prior bypass surgery or stents underscore the need for upstream clinical validation to ensure appropriate application.

\textbf{Ablation Study.}  
To assess the impact of backbone size on segmentation performance, we conducted an ablation study using four variants of the DINOv2 architecture equipped with register tokens: Small, Base, Large, and Giant. Each backbone was paired with the same custom segmentation head and trained on gated CT slices under identical conditions. The evaluation was performed on the Heartlens test set using the same pixel-level metrics as in earlier sections.
\begin{table}[htbp]
\centering
\caption{Ablation results comparing DINOv2 backbone sizes (Small to Giant) on the gated Heartlens test set. Performance metrics include Dice, IoU, Precision, and Recall.}
\label{tab:dinov2_ablation}
\small
\setlength{\tabcolsep}{6pt}
\begin{tabular}{lcccc}
\toprule
\textbf{Backbone} & \textbf{Dice} & \textbf{IoU} & \textbf{Precision} & \textbf{Recall} \\
\midrule
DINOv2-with-registers-Small & 0.672 & 0.651 & 0.667 & 0.675 \\
DINOv2-with-registers-Base  & \textbf{0.865} & \textbf{0.857} & \textbf{0.869} & \textbf{0.867} \\
DINOv2-with-registers-Large & 0.835 & 0.823 & 0.842 & 0.832 \\
DINOv2-with-registers-Giant & 0.801 & 0.788 & 0.814 & 0.790 \\
\bottomrule
\end{tabular}
\end{table}
Results are summarized in Table~\ref{tab:dinov2_ablation}. While performance improved when scaling from Small to Base, further increases in model size led to a decline in segmentation quality. Notably, the Large and Giant models exhibited signs of overfitting, with reduced Dice and IoU scores-likely due to excessive capacity and feature redundancy, which hindered generalization on medical CT data.  
These findings highlight the importance of architectural balance: the Base configuration of DINOv2 delivers the best trade-off between performance and efficiency. As a result, this backbone was adopted in our pipeline for all subsequent experiments.

\section{Discussion and Conclusion}
This study introduced an end-to-end framework for CAC detection and scoring. Its self-supervised \textbf{CARD-ViT} backbone, trained exclusively on gated CTs, achieved strong segmentation and generalized effectively to non-gated scans. These findings demonstrate that self-supervised ViT representations encode robust, transferable anatomical features, enabling reliable, annotation-efficient CAC quantification across scan types using a clinically-aligned, lesion-specific scoring method.

On non-gated CTs, performance was lower than gated evaluation, suggesting that opportunistic use is best framed as triage with clinician over-read rather than fully automated decision-making. Errors were concentrated in intermediate risk ranges (11--400), where modest shifts in peak HU or segmented lesion area near category thresholds can flip categories under motion-related variability.

Key limitations include the lack of pixel-level annotations for non-gated scans, which restricts segmentation evaluation to gated data, and the relatively small non-gated test cohort. Although we predict an artery-agnostic calcium mask, labeled training slices are vessel-imbalanced (LM underrepresented; Table~\ref{tab:dataset_complete}), which may bias learning toward more frequent vessel morphologies; this motivates artery-agnostic outputs to reduce vessel-specific bias. In addition, center-specific ECG-gating and reconstruction may shift motion blur and calcium appearance, motivating multi-center validation. In summary, this framework offers a scalable and annotation-efficient solution for automated CAC analysis. Its strong cross-domain generalization supports its potential for opportunistic cardiovascular risk screening. Future work will focus on 3D extensions, domain adaptation for non-gated performance, vessel-stratified evaluation with imbalance-aware sampling, and multi-center validation across heterogeneous acquisition protocols.
\section{Acknowledgments}
The project described was supported by the University of Kentucky EXCEL Research Initiative through grant 1013170073, and the NIH National Center for Advancing Translational Sciences through grant number UL1TR001998. Additionally, M.S.G. is financially supported through a Study Abroad Program by the Ministry of National Education (Law: 1416), Republic of Turkey. The content is solely the responsibility of the authors and does not necessarily represent the official views of the University, NIH, or the Ministry of National Education.
 
\makeatletter
\renewcommand{\@biblabel}[1]{\hfill #1.}
\makeatother

\bibliographystyle{vancouver}
\bibliography{amia}  
\end{document}